%% file: egpaper_for_review_jcy3.tex
\documentclass[10pt,twocolumn,letterpaper]{article}

\usepackage{iccv}
\usepackage{times}
\usepackage{epsfig}
\usepackage{graphicx}
\usepackage{amsmath}
\usepackage{amssymb}
\usepackage{multirow}

\input{packages}
\input{macros}

\usepackage{graphics}
\makeatletter
  \def\title@font{\Large\bfseries}
  \let\ltx@maketitle\@maketitle
  \def\@maketitle{\bgroup%
    \let\ltx@title\@title%
    \def\@title{\resizebox{\textwidth}{!}{%
      \mbox{\title@font\ltx@title}%
    }}%
    \ltx@maketitle%
  \egroup}
\makeatother

\usepackage[pagebackref=true,breaklinks=true,letterpaper=true,colorlinks,bookmarks=false]{hyperref}

\iccvfinalcopy % *** Uncomment this line for the final submission

\def\iccvPaperID{12456} % *** Enter the ICCV Paper ID here

\ificcvfinal\fi

\title{Highly Personalized Text  Embedding for Image Manipulation by Stable Diffusion}

\author{Inhwa Han\footnote[1]{} ,   Serin Yang\footnote[1]{} ,   Taesung Kwon,   Jong Chul Ye\\
Korea Advanced Institute of Science and Technology (KAIST), Daejeon, South Korea\\
{\tt\small \{inhwahan, yangsr, star.kwon, jong.ye\}@kaist.ac.kr}}

\footnotetext[1]{First two authors contributed equally.}

\begin{document}

\begin{figure*}
\twocolumn[{
\renewcommand\twocolumn[1][]{#1}
\maketitle
    \centerline{\includegraphics[width=1.05\textwidth]{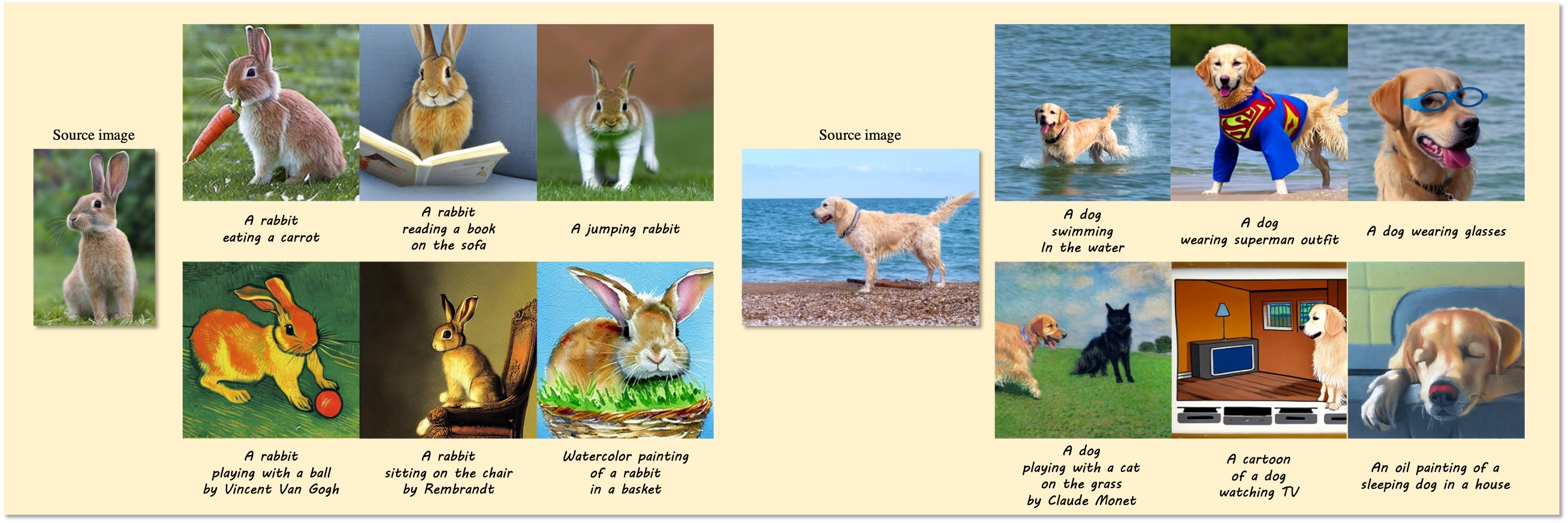}}
        \vspace{-0.2cm}
    \caption{Image manipulation results with highly personalized (HiPer) text embeddings. In the upper row, the identities of the rabbit and the dog are well preserved while adequately manipulating the images to align with target texts. In the bottom row, not only motion and background, but also texture of the source image is transformed towards corresponding target text. }
    \label{fig_main}
        \vspace{0.5cm}
}]
\end{figure*}

\ificcvfinal\thispagestyle{empty}\fi

%%%%%%%%% ABSTRACT
\begin{abstract}
Diffusion models have shown superior performance in image generation and manipulation, but the inherent stochasticity presents challenges in preserving and manipulating image content and identity. While previous approaches like DreamBooth~\cite{ruiz2022dreambooth} and Textual Inversion~\cite{gal2022image} have proposed model or latent representation personalization to maintain the content, their reliance on multiple reference images and complex training limits their practicality. In this paper, we present a simple yet highly effective approach to personalization using highly personalized (HiPer) text embedding by decomposing the CLIP embedding space for personalization and content manipulation.
Our method does not require model fine-tuning or identifiers, yet still enables manipulation of background, texture, and motion with just a single image and target text. Through experiments on diverse target texts, we demonstrate that our approach produces highly personalized and complex semantic image edits across a wide range of tasks. We believe that the novel understanding of the text embedding space presented in this work has the potential to inspire further research across various tasks. Project page is available at 
\href{http://hiper0.github.io/}{http://hiper0.github.io/} 
\end{abstract}

\begin{figure*}[!t]
    \centerline{\includegraphics[width=0.95\textwidth]{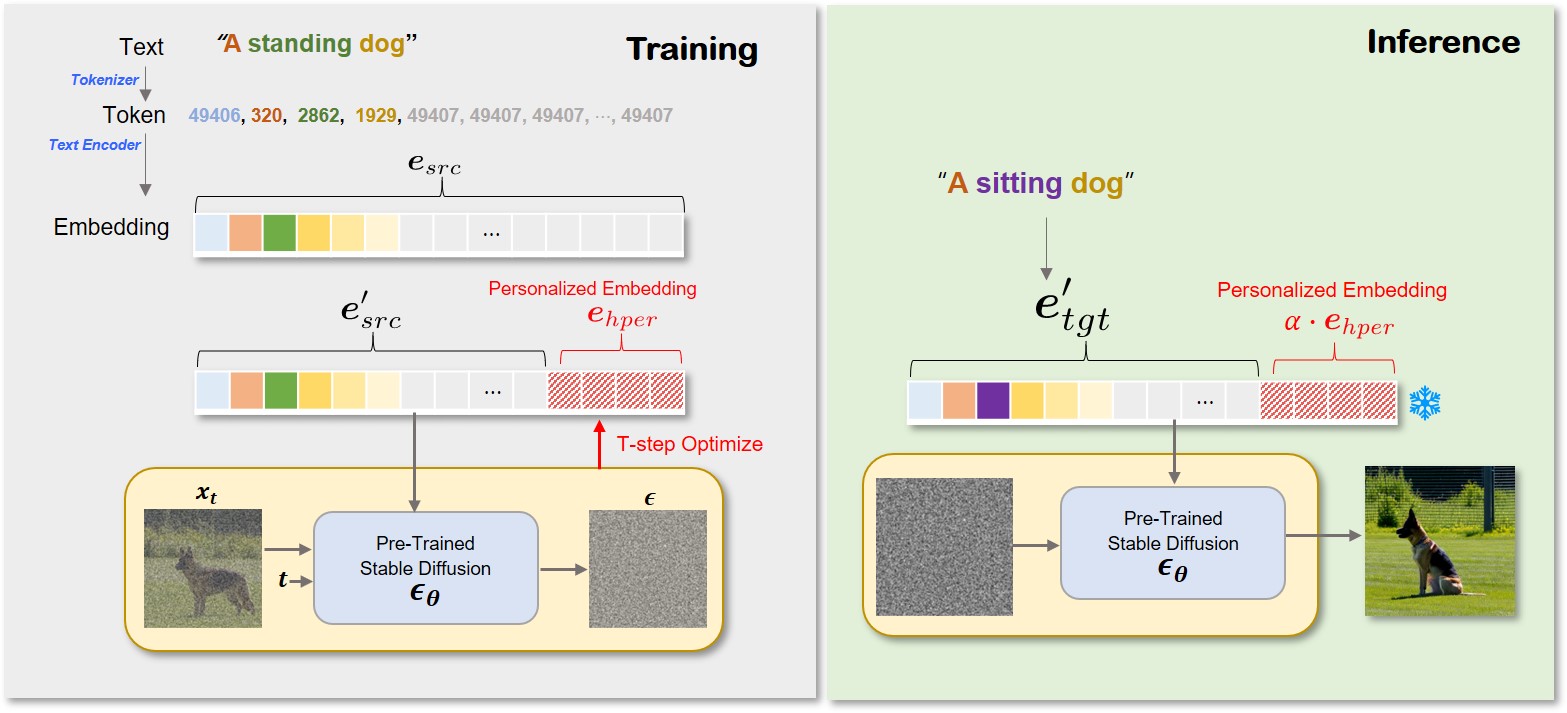}}
    \caption{The proposed method. (Training) First, the source text prompt, which have the meaning of source image, is converted to text embedding. Some parts of text embedding, which have no information, are removed. The informative target embedding part and the personalized embedding is concatenated, and they are the input of pre-trained U-net. In training, the personalized embedding is only optimized. Although this figure depicts it as learning in image space, the embedding is actually optimized in latent space. (Inference) The target embedding is also cropped and concatenated with personalized embedding. Personalized embedding vector is calibrated by multiplying it with $\alpha=0.8$. The pre-trained text-to-image model, which conditioned that embedding, generates an image which has the meaning of target text and the subject of source image.}%
    \label{fig_method}
\end{figure*}

%%%%%%%%% BODY TEXT
\section{Introduction}
% text2image
Image manipulation has long been a subject of interest in image processing. Early works on image manipulation utilized GANs~\cite{shen2020interpreting,harkonen2020ganspace,shen2020interfacegan, dong2017semantic, nam2018text}. However, these approaches required annotated datasets, manual identification of editing directions, or off-the-shelf classifiers.
 The advent of CLIP~\cite{radford2021learning} has made text-guided image manipulation feasible, and combining it with the generative power of StyleGAN~\cite{karras2020analyzing} has led to significant successes with only text prompts, as demonstrated in recent works such as~\cite{Patashnik_2021_ICCV,gal2022stylegan}.

% Diffusion models
Diffusion models have also gained significant attention in the domain of text-guided image synthesis. The inherent stochastic nature of these models has proven to be particularly useful in generating high-quality images from text prompts \cite{avrahami2022blended, kim2022diffusionclip, hertz2022prompt, kwon2022diffusion}.
One notable example of a text-to-image diffusion model is Stable Diffusion \cite{rombach2022high}, which has shown remarkable performance in generating images that align well with the conditioning text prompts. However, despite its impressive results, the stochastic nature of Stable Diffusion can lead to issues with maintaining critical content information, such as the shape or identity of the subject in the generated image. This limitation is not unique to Stable Diffusion but is a fundamental issue with all diffusion models. Therefore, 
researchers have focused on developing effective techniques to ensure that the generated images maintain crucial content information during the conversion of the style and other visual attributes.
%several studies have proposed personalized diffusion models or their text representations to preserve the identity of the subject in an image. 

For instance, DreamBooth \cite{ruiz2022dreambooth} and Textural Inversion \cite{gal2022image} proposed using an identifier to personalize the generative model or text representation, respectively. Although both studies succeeded in preserving the identity of the subject, their application is limited due to the requirement of three to five reference images for optimal performance, which may not be feasible in many scenarios. Additionally, model fine-tuning or optimizing intermediate representation requires a considerable amount of time, which limits their practical use.
Moreover, these prior works have inferior capacity in motion editing, which further restricts their application in scenarios where motion editing is required. 

\begin{figure*}[!t]
    \centerline{\includegraphics[width=0.9\textwidth]{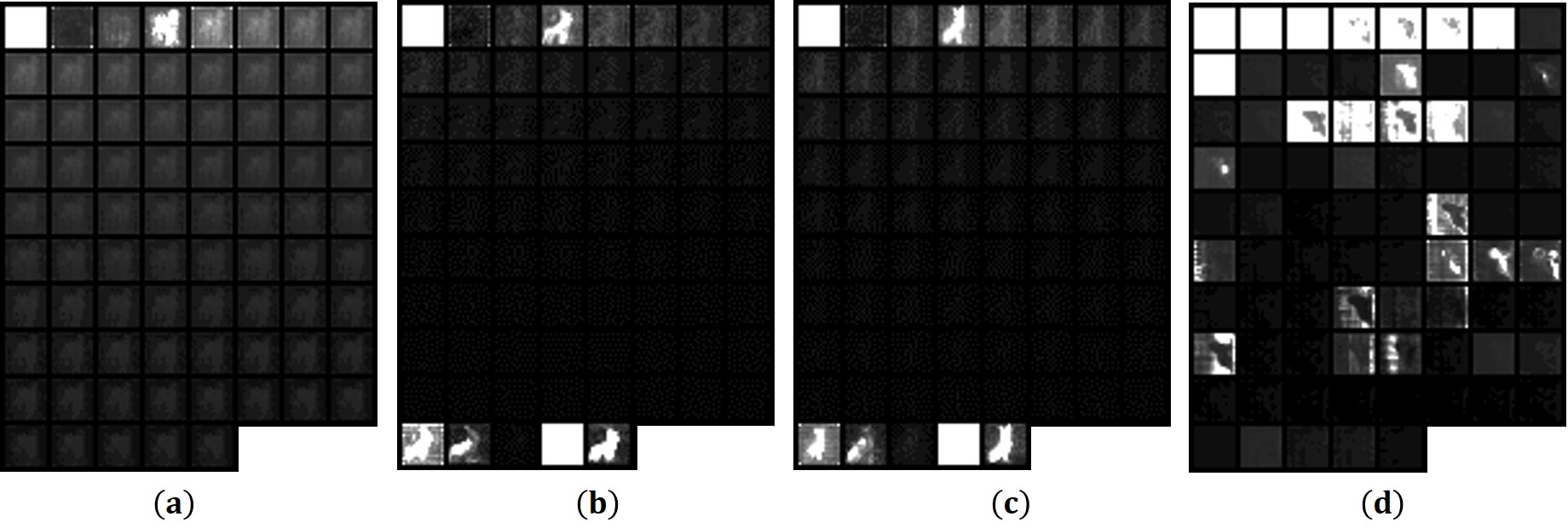}}
    \caption{Cross Attention maps in the final timestep of text-to-image diffusion models. 
    The source text is ``a standing dog' and the target text is ``a sitting dog". Cross Attention maps (a) conditioned with ${\eb}_{src}$ (b) conditioned with $[{\eb}_{src}', {\eb}_{hper}]$, (c) conditioned with $[{\eb}_{tgt}', {\eb}_{hper}]$. (d) Cross attention maps by Imagic \cite{kawar2022imagic} with Stable Diffusion.}
    \label{fig_attention}
\end{figure*}

% Motion editing - Imagic
Imagic \cite{kawar2022imagic} is one of the diffusion-based methods that proposed to solve semantic image editing.
It has shown the potential to edit motion while preserving the underlying structure and background.
However, the method is computationally demanding and involves multiple optimization steps and model fine-tuning, resulting in significant time consumption.
Moreover, the final output is obtained through interpolation of the optimized embedding and CLIP embedding of the target text.
This implies that the output is not an exact embedding aligned with the target text, but rather an approximation in the embedding space.
This makes it hard to maintain the structure information and requires a significant amount of manual work to sample high-quality final outputs.

Therefore, it is imperative to develop more efficient and effective methods that can preserve the identity of the subject while accommodating motion editing and having lower computational costs to make these approaches more practical and applicable in real-world scenarios.

% Personalization - DreamBooth, Textual Inversion

% Ours
In this work, we propose a novel approach to address the limitations of existing diffusion model-based approaches for image manipulation by text prompts. Our method is simple yet highly effective in personalizing Stable Diffusion. One of the key contributions of our work is the discovery of the unique semantic decomposition of the CLIP embedding space in Stable Diffusion \cite{rombach2022high}. We found that while the initial parts of the CLIP embedding space are useful for image manipulation by text tokens, the tail parts play a crucial role in preserving the identity of the subject in the source image. By optimizing the tails, we can retain the identity of the subject while modifying the remaining parts of the image to align with the target prompts. Compared to existing methods such as DreamBooth \cite{ruiz2022dreambooth} and Textural Inversion  \cite{gal2022image}, our approach requires only a single image and target text, without additional training on the diffusion model, resulting in a significant increase in computational efficiency.

To demonstrate the effectiveness of our method, we conducted experiments on diverse target texts with varying backgrounds, textures, and motions. Our results show that our method outperforms existing methods in personalization and semantic manipulation of images by text prompts. Overall, our approach offers a simple yet highly effective solution to the limitations of existing methods and has the potential to advance the field of text-guided image synthesis.
Our contribution can be summarized as following:
\begin{itemize}
\item 
We propose a simple yet highly effective optimization method that does not require model fine-tuning or identifiers, enabling a short training time of approximately 3 minutes.

\item 
We propose a novel method to personalize the text embedding using only a single image. Our approach involves decomposing the CLIP embedding spaces based on their positions. 
\item
Our approach enables manipulation not only of the background or texture but also of motion, which was difficult to achieve using previous methods.
\end{itemize}

%-------------------------------------------------------------------------
\section{Related Work}
\subsection{Text-guided image synthesis}
For text-guided image synthesis using GANs~\cite{li2019object, zhang2018photographic, hinz2020semantic, li2019controllable, xu2018attngan}, text was usually used as condition.
With image-caption data pairs, GANs were trained to generate samples by utilizing attention mechanisms or contrastive approaches~\cite{xu2018attngan,ye2021improving}.
With the introduction of CLIP model~\cite{radford2021learning} which has been trained on a vast dataset of 400 million image-text pairs, the need for annotated images has been reduced.
By combining the great capacity of CLIP and the generative power of StyleGAN~\cite{karras2020analyzing}, works on text-guided image manipulation have been developed with CLIP-driven stylizing losses ~\cite{Patashnik_2021_ICCV,gal2022stylegan}.
Furthermore, CLIPstyler~\cite{kwon2022clipstyler} introduces a CNN encoder-decoder architecture that leverages patch-wise CLIP loss to capture both content and style properties. As a result, the model can generate and manipulate images beyond the domains of pre-trained generators. With autoregressive framework, ~\cite{ramesh2021zero} and ~\cite{ding2021cogview} are text-to-image generators that utilize transformer-based joint pretraining for both vision and language tokens, and have shown superior performance compared to previous GAN-based methods.

\subsection{Diffusion models}
Diffusion models have become popular in the field of generative models for their ability to transform a noise vector into an output image through a diffusion process~\cite{ho2020denoising,song2020denoising,nichol2021improved}. 
Recent research~\cite{dhariwal2021diffusion, song2020score} has demonstrated that the diffusion model exhibits superior image generation quality in comparison to GANs~\cite{brock2018large,zhang2019self}. The diffusion models are recently used for image manipulation task \cite{liu2023more,sasaki2021unit,meng2021sdedit} by adding noise to an image, and then editing the image during the reverse diffusion process. Using the  diffusion models and CLIP model, the text-based image editing methods~\cite{kim2022diffusionclip, avrahami2022blended, nichol2021glide} have been studied, and shown great image manipulation performance. DiffusionCLIP~\cite{kim2022diffusionclip} and Blended Diffusion~\cite{avrahami2022blended} uses CLIP model that regulates the characteristics of the image according to the provided textual prompts during diffusion process. In ~\cite{hertz2022prompt} and ~\cite{parmar2023zero}, the image manipulation methods using diffusion model involves the controlling of cross-attention map.

Beyond image manipulation, text-based image synthesis models using Diffusion models~\cite{saharia2022photorealistic, ramesh2022hierarchical, rombach2022high, nichol2021glide}  has emerged, and generate more accurate image correspond to the given text prompts. Among them, Stable Diffusion~\cite{rombach2022high} is a text-to-image generation model that has attracted attention for its flexibility. It has been recently extended to text-based image manipulation and enables both local and global editing, as well as personalization, allowing for a wide range of image manipulation possibilities. Despite the advantages of diffusion models, they have limitations when it comes to preserving identity information in the generated images.

This has led to significant works on preserving the subject's identity in images by optimizing embeddings or personalizing diffusion models.
Imagic~\cite{kawar2022imagic} proposed a three-step method involving diffusion model fine-tuning to optimize the embedding that generates the exact source image. However, it requires sampling lots of images and may not always produce the desired output, as it interpolates the optimized embedding with a CLIP embedding from the target text in the final step. 
DreamBooth~\cite{ruiz2022dreambooth} utilizes a method that involves using identifiers to preserve the identity of the subject in the source image. While DreamBooth~\cite{ruiz2022dreambooth} can generate images with different conditions while preserving the identity of the subject, it requires multiple images for fine-tuning the diffusion model with complex loss functions, which can be time-consuming and computationally expensive.
Textual Inversion~\cite{gal2022image} finds new pseudo-words by conducting personalization in the text embedding space. These pseudo-words lead to identity preservation. However, Textual Inversion requires multiple images for optimal performance and takes about an hour in order to optimize the embedding vector associated with the pseudo-word.

%------------------------------------------------------------------------
\section{Highly Personalized Text Embedding (HiPer)}

\subsection{Stable Diffusion Model}
 
  Diffusion models~\cite{ho2020denoising} attempt to model the data distribution $p_{\rm data}(\xb)$ by constructing a hierarchical latent variable model.
 Specifically, % in diffusion models, given a source data distribution $q(x_0)$, latent variable $x_t$ is computed by forward diffusion process. 
DDPM~\cite{ho2020denoising} directly samples $\x_t$ from a clean image $\x_0$ by adding Gaussian noise with $\beta_t \in (0, 1)$ at time $t \in [1,...,T]$,
\begin{equation} \label{eq_ddpm_forward}
\x_t = \sqrt{\bar{\alpha}_t}\x_0 + \sqrt{1-\bar{\alpha}_t}\epsilonb
\end{equation}
where $\epsilonb \sim \mathcal{N}(0,\I)$, $\alpha_t = 1 - \beta_t$, and $\bar{\alpha}_t = \prod_{i=0}^{t}\alpha_i$. The reverse sampling process to generate a clean image is then given by:

\begin{equation} \label{eq_ddpm_reverse}
\x_{t-1} 
= \frac{1}{\sqrt{1-\beta_t}}
\left(\x_t - \frac{\beta_t}{\sqrt{1-\overline{\alpha}_t}}\epsilonb_{\theta}(\x_t,t)\right)
+\sigma_{t}\epsilonb.
\end{equation}
where 
where $\sigma_t := \frac{1 - \bar\alpha_{t-1}}{1 - \bar\alpha_t}\beta_t$ and
the neural network $\epsilonb_{\theta}(x_t,t)$ 
is trained by 
%is used to estimate the noise component, which can be viewed as a score function up to a scaling factor.
%Training of diffusion models amount to training a multi-noise level residual denoiser (i.e. epsilon matching)
\begin{align*}
    \min_\theta \Ed_{\xb_t, \xb_0, \epsilonb \sim \Nc(0, \Ib)}\left[\|\epsilonb_\theta(\xb_t,t) - \epsilonb\|_2^2\right],
\end{align*}

Stable Diffusion Models (SDM) \cite{rombach2022high} is a publicly available diffusion model with important modifications. Instead of using diffusion sampling directly in the image space, SDM is based on the latent diffusion method, where the forward and reverse diffusion sampling in  \eqref{eq_ddpm_forward} and \eqref{eq_ddpm_reverse} are performed in the latent domain. To achieve this, SDM uses a VQ-GAN \cite{esser2021taming} encoder $E$ to convert the image $\pb$ to a lower-dimensional latent variable $\xb$ (i.e., $\xb = E(\pb)$). The diffusion sampling is then performed in the latent space, and the final image is obtained using a VQ-GAN decoder $D$.
Another important modification in SDM is the addition of text-based conditioning. Specifically, SDM uses the tokenizer of the CLIP text encoder ${\tau_\phi}$ to extract words from a given text prompt $\yb$ and convert them into numbers called tokens, denoted by $\eb=\tau_\phi(\yb)$. The tokens are then transformed into text embeddings, which are used to condition the neural network during training:
\begin{align*}
    \min_\theta \Ed_{\xb_t, \xb_0, \epsilonb \sim \Nc(0, \Ib)}\left[\|\epsilonb_\theta(\xb_t,t,\tau_\phi(\yb)) - \epsilonb\|_2^2\right],
\end{align*}
Therefore, SDM enables text-driven image generation using the reverse diffusion sampling in  \eqref{eq_ddpm_reverse} in the latent space. Instead of using $\epsilon_\theta(\x_t,t)$, the model uses a text-conditioned neural network called $\epsilonb_\theta(\xb_t,t,\tau_\phi(\yb))$.

The proposed method is implemented using pretrained Stable Diffusion model, which is publicly available.

\begin{figure*}[!t]
    \centerline{\includegraphics[width=1.0  \textwidth]{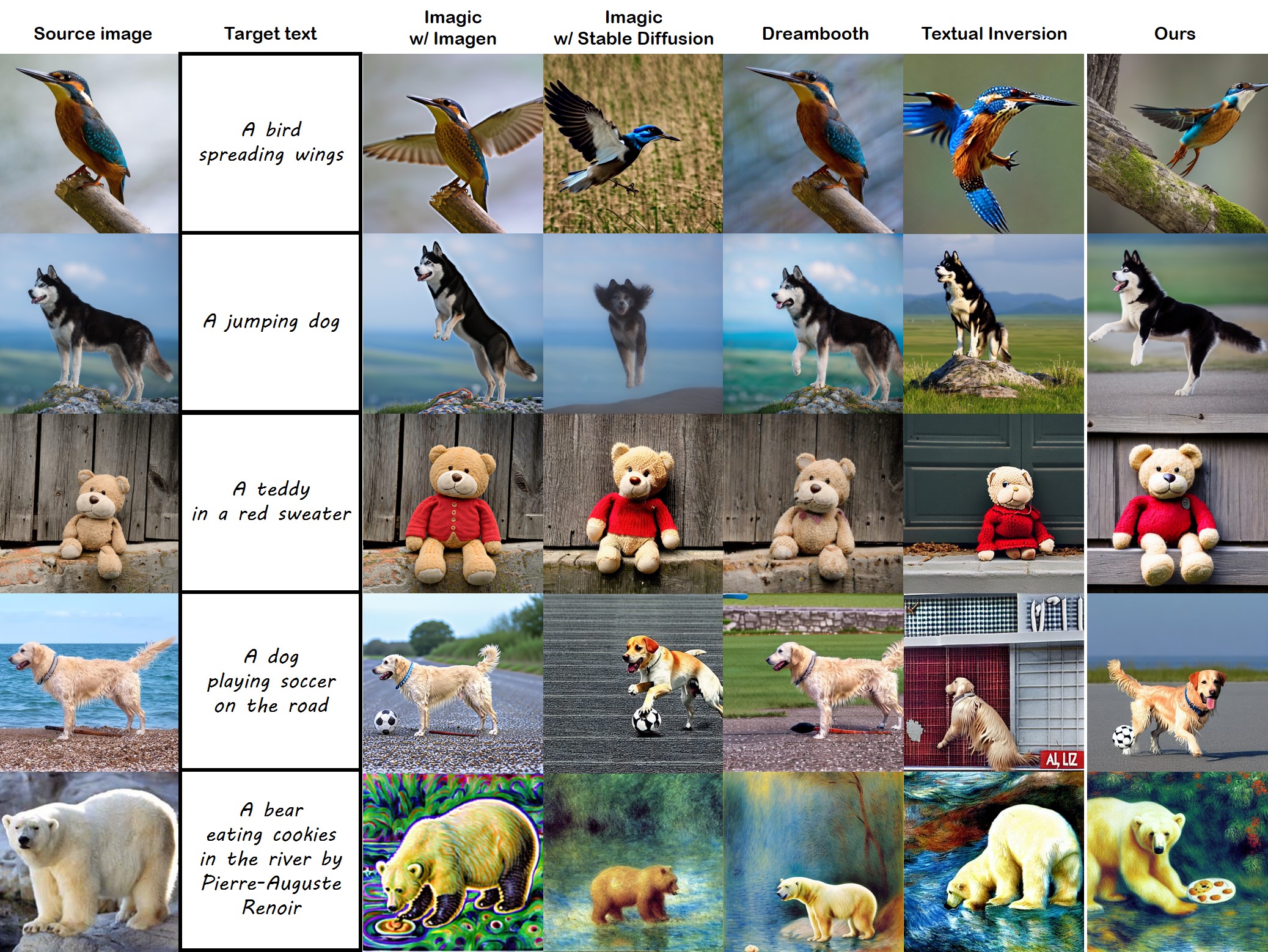}}
    \caption{The qualitative comparison results. Compared with three stable diffusion-based text-guided image manipulation methods, our method shows its superiority. It could preserve the identities of the subject in source images, while appropriately transforming the semantic information to align with the CLIP embedding of the target text.
    Original Imagic results with Imagen show comparable results by using
    proprietary text embedding scheme.}
    \vspace{0.5cm}
    \label{fig_comparison}
\end{figure*}

\begin{figure*}[!t]
    \centerline{\includegraphics[width=\textwidth]{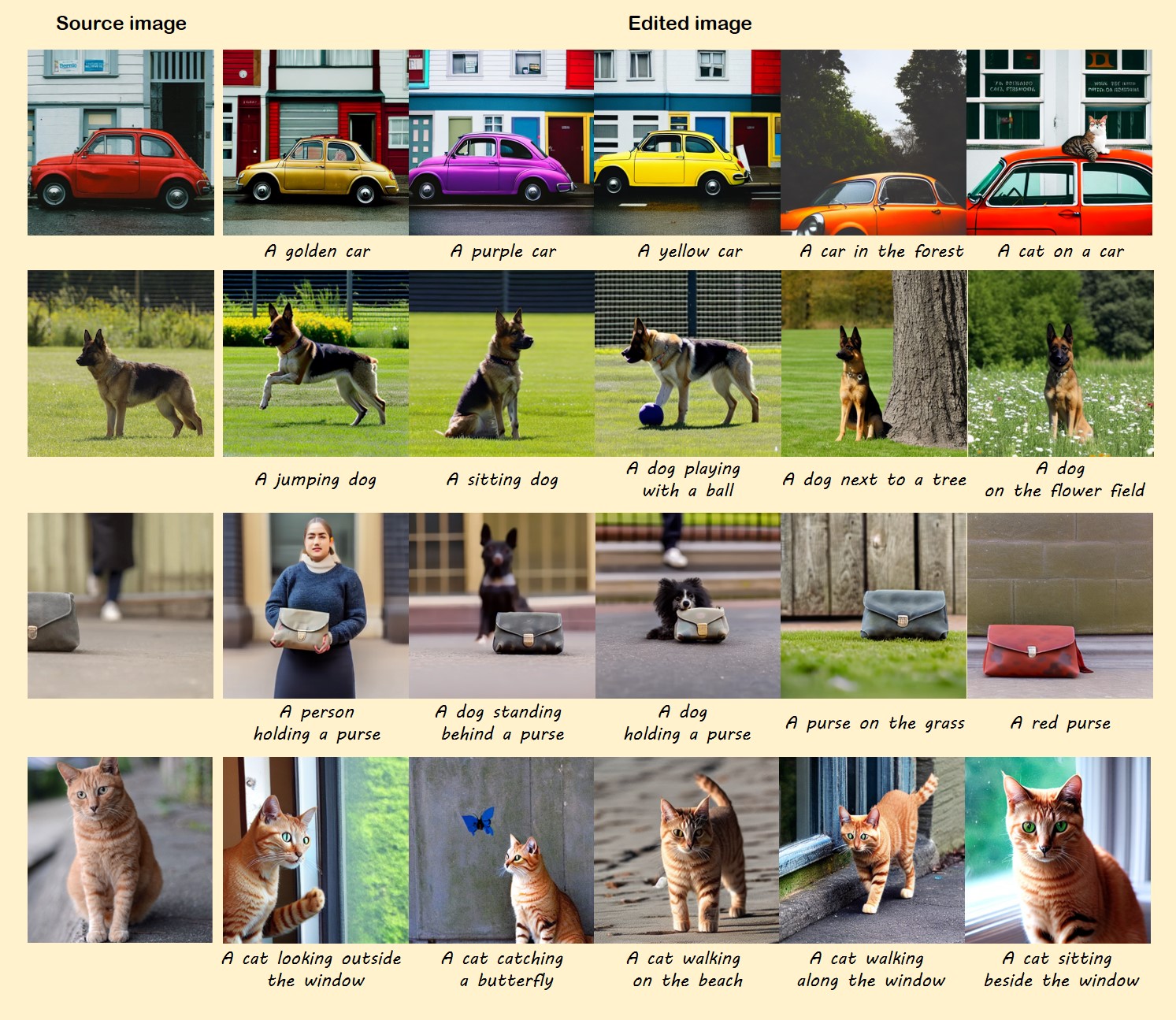}}
    \caption{By concatenating highly personalized (HiPer) text embeddings with different target embeddings, we can achieve precise image manipulation results. This allows us to manipulate the image with high precision while preserving the subject's identity in the source image.}
                \vspace{-0.2cm}
    \label{fig_results}
\end{figure*}

\subsection{Text Embedding Decomposition}

\paragraph{Key observation}
The objective of the proposed method is to optimize text embeddings to better represent a given subject. 
 In SDM, the input text is first converted into an embedding by a tokenizer and text encoder, which is then conditioned to autoencoder. Let ${\eb}_{src}$ be the text embedding for source image. The converted text embedding contains the information of the text. For example, consider an image that corresponds to the text prompt ``a standing dog". The source embedding has the information of ``a standing dog". However, if the length of the input text is shorter than the maximum length of the text embedding, then the end part of the embedding does not contain any information about the input text (See Fig.~\ref{fig_attention}(a)).
 Taking this into account, the proposed method only optimizes a piece of the text embedding at the end, which has no information about the source text. If successfully optimized, the piece of embedding would contain a lot of information about the source image.
 
 \paragraph{HiPer embedding}
Consider a source text embedding $\eb_{src} \in \Rd^{C\times M}$  for a text prompt $\yb$, i.e. $\eb_{src}=\tau_\phi(\yb)$ (See Figure~\ref{fig_method}).
 In the proposed method, we select the last $N$ tokens at the end of the text as what we call the highly personalized (HiPer) embedding ${\eb}_{hper}\in \mathbb{R}^{C\times N}$ for image representation, and the remaining $M-N$ tokens is denoted by  ${\eb}_{sem}'\in \mathbb{R}^{C\times (M-N)}$. More specifically, this leads to the following decomposition of the text embedding:
 \begin{align}
 \eb_{src} = \begin{bmatrix} \eb'_{src}, \eb_{hper} \end{bmatrix}
 \end{align}
 Since the uninformative part is removed, ${\eb}_{src}'$ still contains the information of source text similar to ${\eb}_{src}$. 
 Then,  for a given image $\pb_0$ and its latent $\xb_0=E(\pb_0)$, 
the HiPer embedding $\eb_{hper}$ is only optimized whereas $\eb_{src}'$  generated from the source text is maintained:
\begin{align} \label{eq_}
&\eb_{hper} \notag\\
&= \arg\min\limits_{\eb_{h}\in \mathbb{R}^{C\times N}} \mathbb{E}_{\x_t, \epsilonb\mathtt{\sim}\mathcal{N}(0,\Ib)}\left[\|\epsilonb- \epsilonb_{\theta}(\x_t,t,[{\eb}_{src}', {\eb}_{h}])\|^2\right] 
\end{align}
Note that this is different from Textual Inversion \cite{gal2022image}, where the whole text embedding $\eb_{src}$ is optimized:
\begin{align} %\label{eq_}
\eb_{src} = \arg\min\limits_{\eb\in \mathbb{R}^{C\times M}} \mathbb{E}_{\x_t,\epsilonb\mathtt{\sim}\mathcal{N}(0,\Ib)}\left[\|\epsilonb- \epsilonb_{\theta}(\x_t,t,\eb)\|^2\right] 
\end{align}

In the inference step, ${\eb}_{src}'$ is replaced with the cropped text embedding ${\eb}_{tgt}'$ of the target
prompt, while the tail $N$ tokens are replaced with the HiPer embedding $\eb_{hper}$. 
Then, the reverse diffusion generates latent samples by
\begin{equation} %\label{eq_ddpm_reverse}
\x_{t-1} 
= \frac{1}{\sqrt{1-\beta_t}}
\left(\x_t - \frac{\beta_t}{\sqrt{1-\overline{\alpha}_t}}\epsilonb_{\theta}(\x_t,t,\eb_{cmp})\right)
+\sigma_{t}\epsilonb.
\end{equation}
where the composite embedding $\eb_{cmp}$ is given by $$\eb_{cmp}= [{\eb}_{tgt}', {\eb}_{hper}] .$$
Subsequently, the final image is obtained by applying decoder to the final latent sample, i.e. $\x_0$.

When the composite embedding $\eb_{cmp}$ is conditioned to pre-trained Stable Diffusion model, we found that the output image has the meaning of target text and the personalization from the subject of the source image. For instance, if the source image has the meaning of ``a standing dog" and the target text is ``a sitting dog", the target embedding with personalized embedding generates the source image's dog, which is sitting. The simplicity of our method contrasts with its impressive performance results, which will be demonstrated in the following section.

\begin{figure}[!t]
    \centerline{\includegraphics[width=0.5\textwidth]{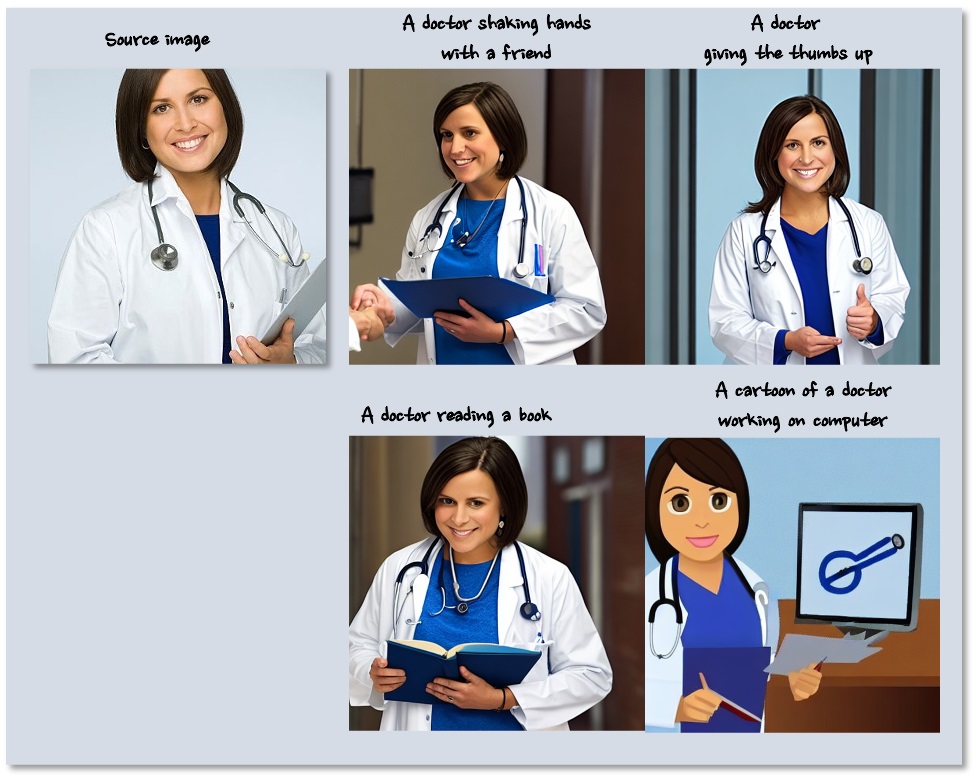}}
    \caption{Text-driven image manipulation results featuring a female doctor.}
    \vspace{-0.5cm}
    \label{fig_human}
\end{figure}

\begin{figure*}[!t]
    \centerline{\includegraphics[width=\textwidth]{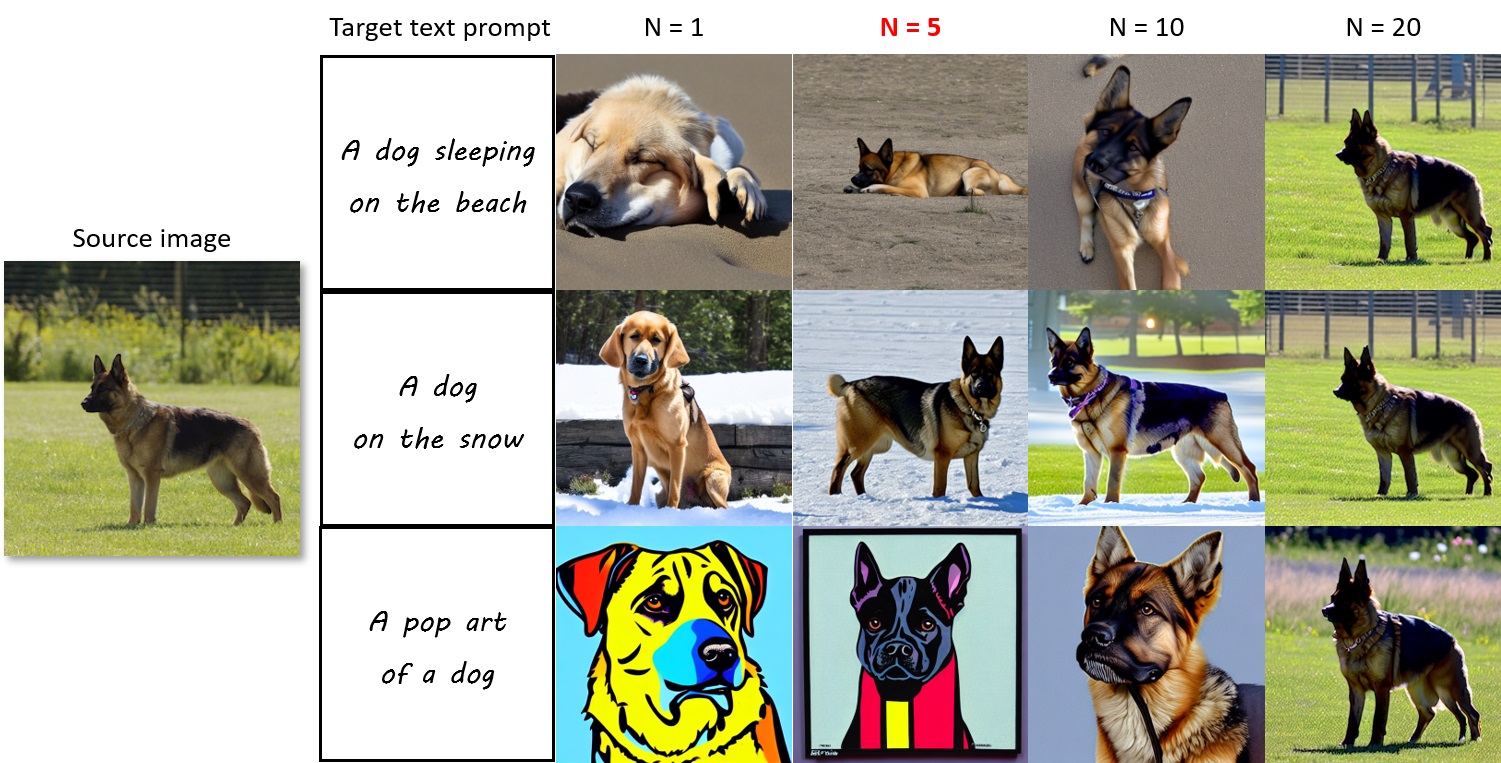}}
    \caption{The effect of the number of personalized tokens, $N$. Increasing $N$
produces outputs similar to the source image. On the other hand, 
smaller values of $N$ generate adequately edited outputs but 
do not sufficiently preserve the identity.}
    \vspace{-0.3cm}
    \label{fig_ntokens}
\end{figure*}

%-------------------------------------------------------------------------
\section{Experimental Results}
Our method is based on Stable Diffusion~\cite{rombach2022high}.
We set the number of optimization steps as $1000$ and the number of personalized tokens $N=5$. The dataset used as source images are from Ted of Imagic~\cite{kawar2022imagic} and LAION~\cite{schuhmann2022laion}. We experimented with the models using PyTorch library on NVIDIA GeForce GTX 3090.

\paragraph{Text-driven image manipulation}
As shown in Figure \ref{fig_main}, our method could achieve image manipulation in all the three areas - motion, background, and texture. The rabbit and the dog in the source images could eat, read books, jump, and swim in diverse backgrounds. In addition to motion and background, texture of images could be controlled. The real source images were well translated into diverse painting styles. Furthermore, the subject identity is well preserved in the generated images. The shape of the rabbit's ears and the color of its fur are maintained in the generated images. Also, the dog's identity, such as the color, species, and fur texture, was contained throughout the output images. Even the appearance of the dog leash is also preserved. 

Moreover, our method exhibits impressive capability in simultaneously manipulating three areas, which are typically challenging to achieve. In Figure \ref{fig_main}, the motion ``playing with a cat," the background ``on the grass," and the texture ``Claude Monet" style are appropriately modulated while preserving the identity of the dog in the source image. 

\paragraph{Comparative studies}
We compared our model with three diffusion model based image manipulation baselines, i.e. Imagic~\cite{kawar2022imagic}, DreamBooth~\cite{ruiz2022dreambooth}, and Textual Inversion~\cite{gal2022image}. Although DreamBooth and Textual Inversion require three to five images for optimal training, we trained them with a single image because of their unavailability. We compared our method with these three methods which were implemented based on Stable Diffusion~\cite{rombach2022high}. Additionally, the Imagic authors provided us with the official Imagic results that were obtained using Imagen, as a model that is not publicly available. We also included them for comparison. The qualitative results are illustrated in Figure \ref{fig_comparison}. 

It is noticed from the results that the implementation of Imagic using the Stable Diffusion model exhibits inferior content preservation capabilities when compared to the official Imagic result and the other results. Although Imagic was proposed to edit semantic information while preserving the other overall shapes, it showed inferior performance with Stable Diffusion model. In the first row, the bird's head is different from the source image and the dog in the second row has lots of hair around its head in the generated image which is different from the source image. It is due to interpolation in the inference step, which approximates between the target and the optimized embedding. The Imagen model-based implementation of Imagic yields significantly improved image manipulation results, preserving the original source image while making targeted edits based on specified textual inputs. However, due to the Imagic model's strong tendency to preserve the object's position and background of the source image, it results in poor performance when the source image undergoes significant changes due to target textual inputs. As shown in the last row of the figure, when the white bear in source image undergoes substantial transformations due to textual inputs, the model fails to preserve its identity.

In case of DreamBooth, the diffusion model is fine-tuned on a single image which results in the output image same as the source one. The semantic editing along the target text is not adequately applied. On the other hand, Textual Inversion shows more plausible results compared to DreamBooth. However, we observed in the fourth and the last rows of Figure \ref{fig_comparison} that its performance degraded when dealing with simultaneous transformation of motion, background, and texture. The dog in the generated image plays with a ball but not on the road. The ball also does not look like a soccer ball. Also, cookies cannot be seen from the output images in the last row. 

Our proposed method overcomes the limitations of existing diffusion-based methods in editing multiple aspects and is able to generate images that maintain the same identity as the source while effectively transforming the attributes to align with the target text. Notably, our method enables a high level of manipulation, as evidenced by the bear in the last row, which is shown eating cookies in the river, a transformation that was not possible using any existing method.

\paragraph{More examples}
In Figure~\ref{fig_results}, we present the impressive text-driven image manipulation results achieved through our method. By leveraging highly personalized (HiPer) text embeddings and concatenating them with different target embeddings, we are able to achieve precise image manipulation without compromising the subject's identity in the source image. Our method empowers users to control various visual attributes of the image, such as colors, movement, background, and detailed scene description, while preserving the original image's integrity. This not only fosters creative expression but also has numerous potential applications in fields such as fashion, advertising, and entertainment. Moreover, our method is computationally efficient and can be applied to a diverse range of images, making it a practical and versatile tool for image manipulation.

Figure~\ref{fig_human} showcases the text-driven image manipulation results featuring a doctor. The original image depicts the doctor wearing a white lab coat and standing against a plain background. The manipulation results highlight our method's ability to modify specific visual attributes of the image based on the input text while maintaining the integrity of the original image. In this case, the doctor's posture and hand gesture were modified based on the input text, resulting in an image where the doctor appears to be giving the thumbs up. Additionally, the image was converted to a cartoon style, demonstrating our method's versatility in handling different image styles. Overall, this example illustrates the potential of our method in various applications.

Additional and more comprehensive experimental results can be found in the Supplementary Materials.

\section{Discussion}
\paragraph{Number of personalized tokens}
We performed experiments with varying numbers of tokens $N$ to study their effect on personalization. Figure \ref{fig_ntokens} illustrates that increasing the number of tokens leads to overfitting of the personalized embedding, producing outputs similar to the source image and failing to capture the desired changes. In the last column, the dogs in the generated image are not shown as sleeping, the background does not feature snow, and the texture has not been transformed into a pop art style. On the other hand, smaller values of $N$ generate adequately edited outputs but do not sufficiently preserve the identity. It is shown that the dogs generated with $N=1$ are different from the source image. Based on these findings, we chose $N=5$ as our baseline.

\paragraph{Comparison of cross-attention maps}
Our method is based on a key observation that the cross-attention map in text-to-image diffusion models does not contain any information at the end of the token.
 Figure \ref{fig_attention} show the cross-attention map in the final timestep of text-to-image diffusion model conditioned with other embedding. The total number of cross attention map is 77, which corresponds to the total number of token. The source image is the input image of Fig. \ref{fig_method}, which correspond the source text of ``a standing dog', and source embedding of ${\eb}_{src}$, The target text is ``a sitting dog" and the corresponding embedding is ${\eb}_{tgt}$. With the proposed method, the personalized embedding ${\eb}_{hper}$ is optimized. Each represents cross-attention map (a) conditioned with ${\eb}_{src}$, (b) conditioned with $[{\eb}_{src}', {\eb}_{hper}]$, (c) conditioned with $[{\eb}_{tgt}', {\eb}_{hper}]$. 
 Figure \ref{fig_attention} (a) of ${\eb}_{src}$ shows that the attention map corresponding to each text is only activated, and the end of the token is not activated. (b) and (c) shows that the map related to personalized embedding is activated. In the case of target embedding with personalized embedding (c), some attention map related the personalized embedding follows the target embedding part's attention map, but maintain the distribution of personalized embedding, compared to (b). 
In Figure~\ref{fig_attention}(d), the attention map of Imagic with Stable Diffusion is shown, revealing no discernible organized attention pattern.
 
The results confirm that our composite embedding approach, utilizing a highly personalized embedding and a semantic embedding for content, successfully separates the personalization and manipulation aspects. This approach outperforms existing methods which do not separate these two aspects, as demonstrated by our experimental results.

\section{Conclusion}
We present a highly personalized text-to-image generation method using Stable Diffusion, which is simple yet powerful. 
With just a single image, our method can produce highly personalized text tokens, resulting in superior performance in maintaining the identity of the subject. 
Furthermore, our approach does not necessitate model fine-tuning or complex loss functions. 
These properties enable us to manipulate images quickly and easily using a simple optimization process that takes just three minutes. 
Additionally, we have demonstrated the remarkable capabilities of our method by demonstrating image editing results in three areas: motion, background, and texture.

{\small
\bibliographystyle{ieee_fullname}
\bibliography{egbib}
}

\clearpage

\appendix
{\noindent \Large \bf {Supplementary Material}}

\section{Details of experimental setting }
\paragraph{Implementation details} 
For optimizing the personalized embedding with Stable Diffusion model, we set the size of source image to $512\times512$. We use the highly personalized (HiPer) embedding which is optimized during 1000 steps, with the learning rate of $5\times{10}^{-3}$.

\paragraph{The number of HiPer embedding tokens} 
The default number of HiPer embedding tokens was 5. In cases where the generated image is intended to more closely follow the original image, the number of tokens can be increased.

\paragraph{Calibration of HiPer embedding} 
Before concatenating the HiPer embedding  $\eb_{hper}$ with  target embedding $\eb_{tgt}'$, the  HiPer embedding vector is calibrated by multiplying it with 0.8. This value was obtained through experimentation and is a constant value throughout the experiment.

\section{Evaluation}
\subsection{Quantitative analysis}
\begin{figure*}[!t]
    %\vspace{1.3cm}
    \centerline{\includegraphics[width=1.0 \textwidth]{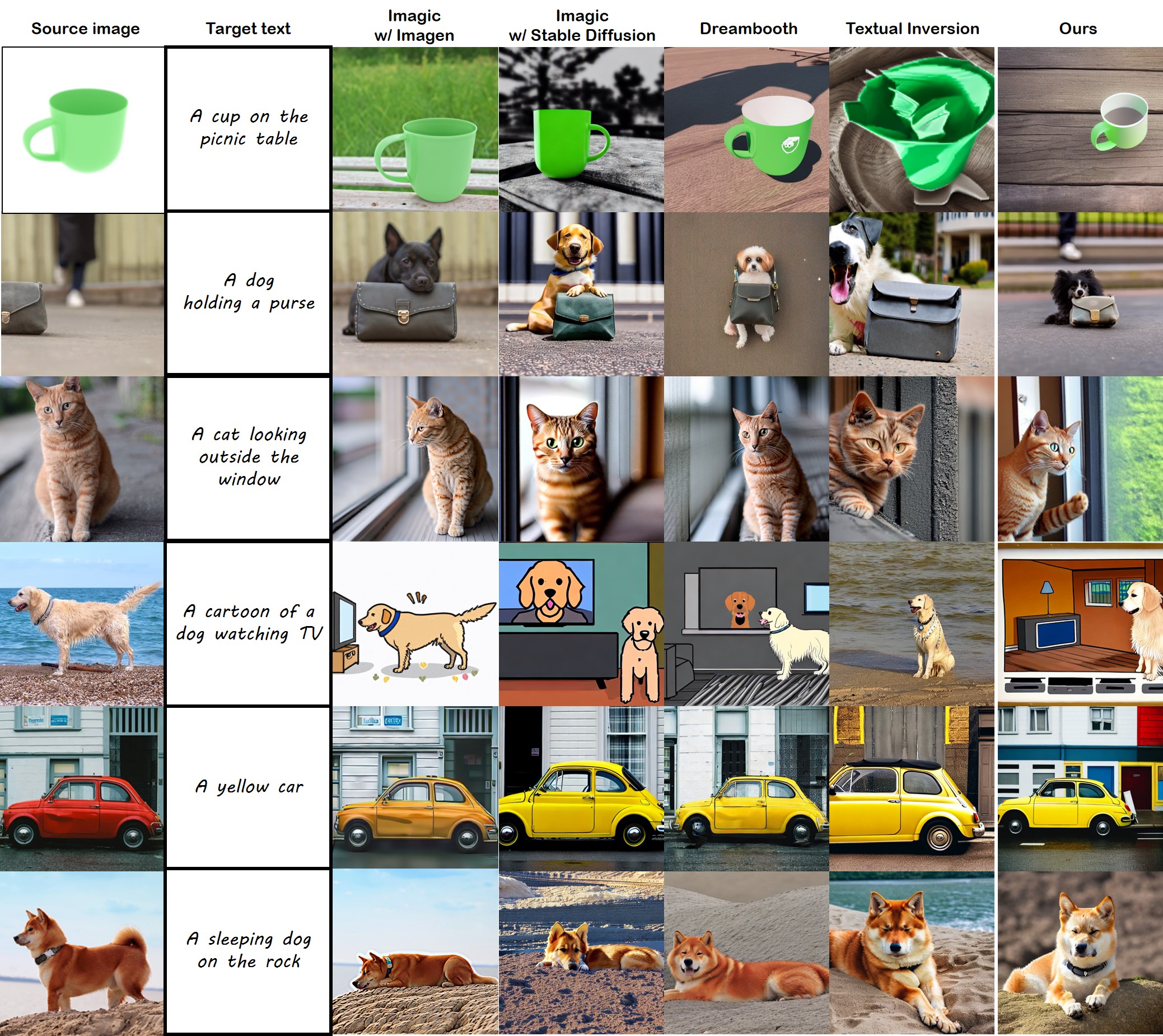}}
    \caption{%The generated images from same text prompt with the proposed method and the other methods.
    The images generated from the same text prompt using our proposed method and other methods are shown. As illustrated, our method consistently produces images that are visually more aligned with the given text prompts, and are more similar to the source images in terms of identity. The images generated by the other methods lack detail and often produce unrealistic results.}
    \vspace{-0.3cm}
    \label{sup_control}
\end{figure*}

For quantitative comparison, we conducted a user study and calculated CLIP scores. The details of the user study will be described later.
To measure the preservation of identity, we calculated the distance between the output images and the source image. In addition, we evaluated the semantic alignment of the output images with the target prompts by computing the distance between them in the CLIP embedding space. As illustrated in Table \ref{table_comparison}, which displays the
average scores from the experiments in Figs. \ref{fig_comparison} and \ref{sup_control},   our method achieved best score in the user study on semantic alignment. While DreamBooth ranked first in the user study for identity preservation, its performance in manipulation was inferior. Its strong focus on identity preservation resulted in inadequate editing towards the target prompt. As shown in Figure \ref{fig_comparison}, the bird in the first row was not transformed enough to align with target text prompt. Meanwhile, Imagic with Stable diffusion showed the good performance in manipulation (cf. CLIP score). However, its output images could not preserve the identities. Figure \ref{sup_control} demonstrates that the shape of the purse in the second row and the dog in the last row differ from the source image.

\begin{table}[hbt]
%\vspace{0.3cm}
\centering
\resizebox{\columnwidth}{!}{
\begin{tabular}{c|p{1.5cm}p{1.5cm}|p{2cm}p{2cm}|p{2cm}}
    \hline
    \rule{0pt}{1\normalbaselineskip}
    Methods 
    &\multicolumn{2}{c}{User study}  
    &\multicolumn{2}{|c|}{CLIP score}   
    &\multicolumn{1}{c}{\multirow{2}{*}{Train time}} \\ [0.5ex]
    w/ Stable Diffusion
    &\multicolumn{1}{c}{Semantic $\uparrow$}  
    &\multicolumn{1}{c}{Identity $\uparrow$}
    &\multicolumn{1}{|c}{Text $\uparrow$}
    &\multicolumn{1}{c|}{Image $\uparrow$}  \\[0.5ex]
    \hline

    \rule{0pt}{1\normalbaselineskip}
    Imagic w/ Stable Diffusion             &\hfil \underline{3.731}   &\hfil 3.251                    &\hfil \bf{0.2148} &\hfil  0.8223    &\hfil 14.08 min\\
    DreamBooth          &\hfil 3.099        &\hfil \bf{4.503}   &\hfil 0.1955                  &\hfil \bf{0.8520}   &\hfil 14.67 min\\
    Textual Inversion   &\hfil 3.070        &\hfil 3.567                    &\hfil 0.1633                  &\hfil 0.8303    &\hfil 42.8 min\\
    HiPer               &\hfil \bf{4.520}   &\hfil \underline{4.099}   &\hfil \underline{0.2047}             &\hfil \underline{0.8443}  &\hfil 3 min\\[0.5ex]
    \hline
\end{tabular}
}
\caption{Quantitative results on comparative studies. The bold text denotes the best score, and the underline indicates the second-best score. All of the models were implemented using the Stable Diffusion model.}
\label{table_comparison}
\end{table}

\begin{figure*}[!t]
    \centerline{\includegraphics[width=1.95\columnwidth]{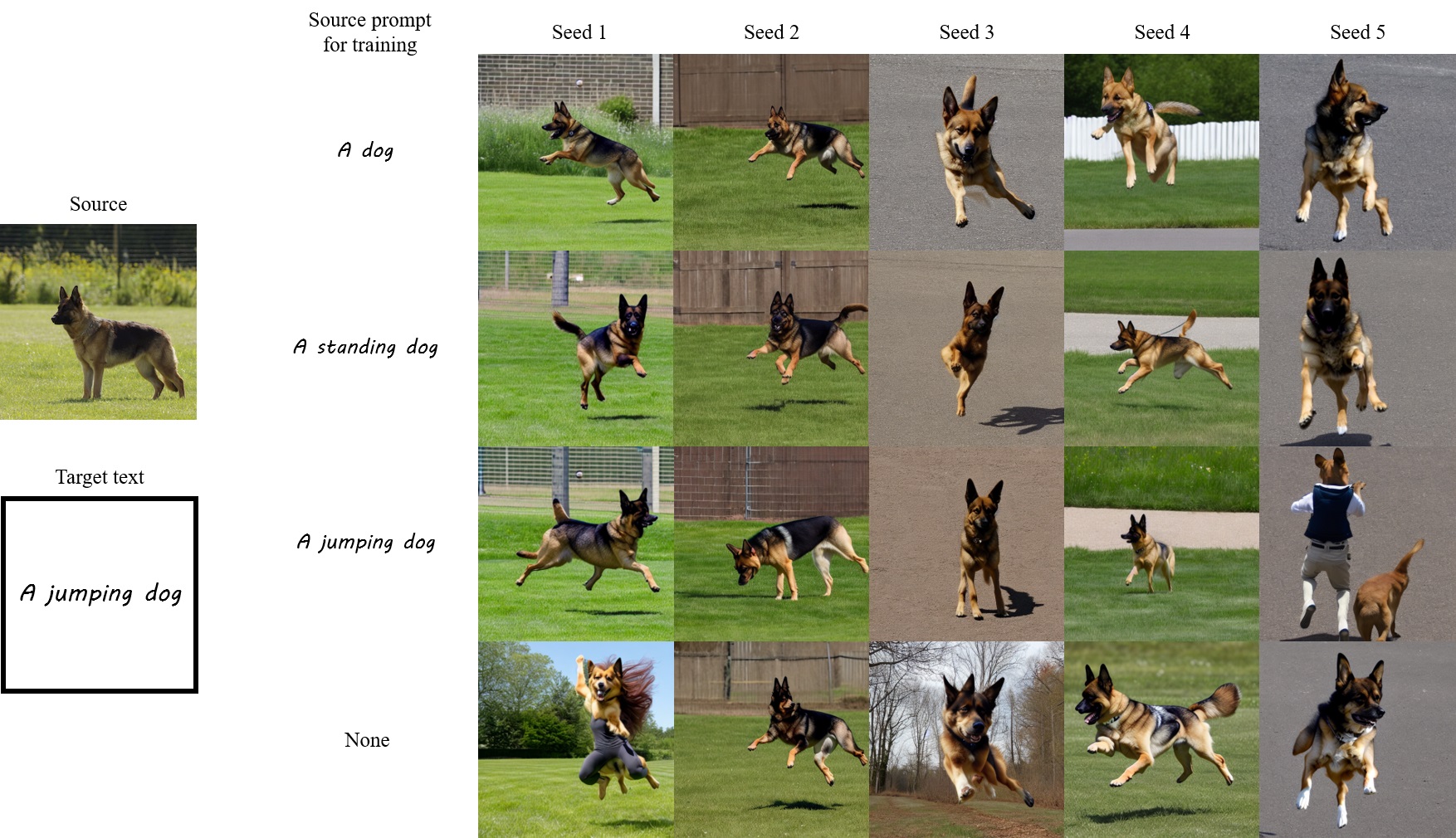}}
    \caption{Ablation study on the source embeddings.}%
    \label{fig_ablation_esrc}
\end{figure*}

\paragraph{User Study}
We conducted a user study to evaluate the performance of our personalized image generation methods, Imagic~\cite{kawar2022imagic}, Dreambooth~\cite{ruiz2022dreambooth}, and Textual Inversion~\cite{gal2022image}, on various tasks such as motion, background, and texture changes. To compare our methods, we used a single source image and conducted the experiments on publicly available Stable Diffusion models. We obtained feedback from 20 participants and asked them to rank the resulting images from other models based on their similarity to the source image (Identity) and their alignment with the given text (Semantic alignment). To gather comprehensive feedback, we utilized a personalized opinion evaluation system through Google Forms, which included a scoring scale of 1-5, with 1 representing the lowest score and 5 representing the highest score. Respondents had the option of selecting one of five different scores for each inquiry: 1-very low, 2-low, 3-middle, 4-high, 5-very high.

%We conduct a user study to evaluate the personalization performance on various tasks such as motion/background/texture change with our method, Imagic~\cite{kawar2022imagic}, Dreambooth~\cite{ruiz2022dreambooth}, and Textual Inversion~\cite{gal2022image}. For comparison, we used only a sinlge image for source image, and conducted the experiment on Stable Diffusion models, which publicly available. We get the answer from 20 people, and we ask the respondents to rank the result image from other models in the respects of 'how the object of result image is similar to that of source image'(Identity), and 'how the result image has the meaning of the given text'(Semantic alignment). To obtain comprehensive feedback from users, we employed a personalized opinion evaluation system using Google Form. The system's scoring scale ranged from 1 to 5, with 1 representing the minimum score and 5 representing the maximum. For each inquiry, respondents had the option of selecting one of five different scores: 1-very low, 2-low, 3-middle, 4-high, 5-very high.

\begin{figure}[!t]
    \centerline{\includegraphics[width=1.05\columnwidth]{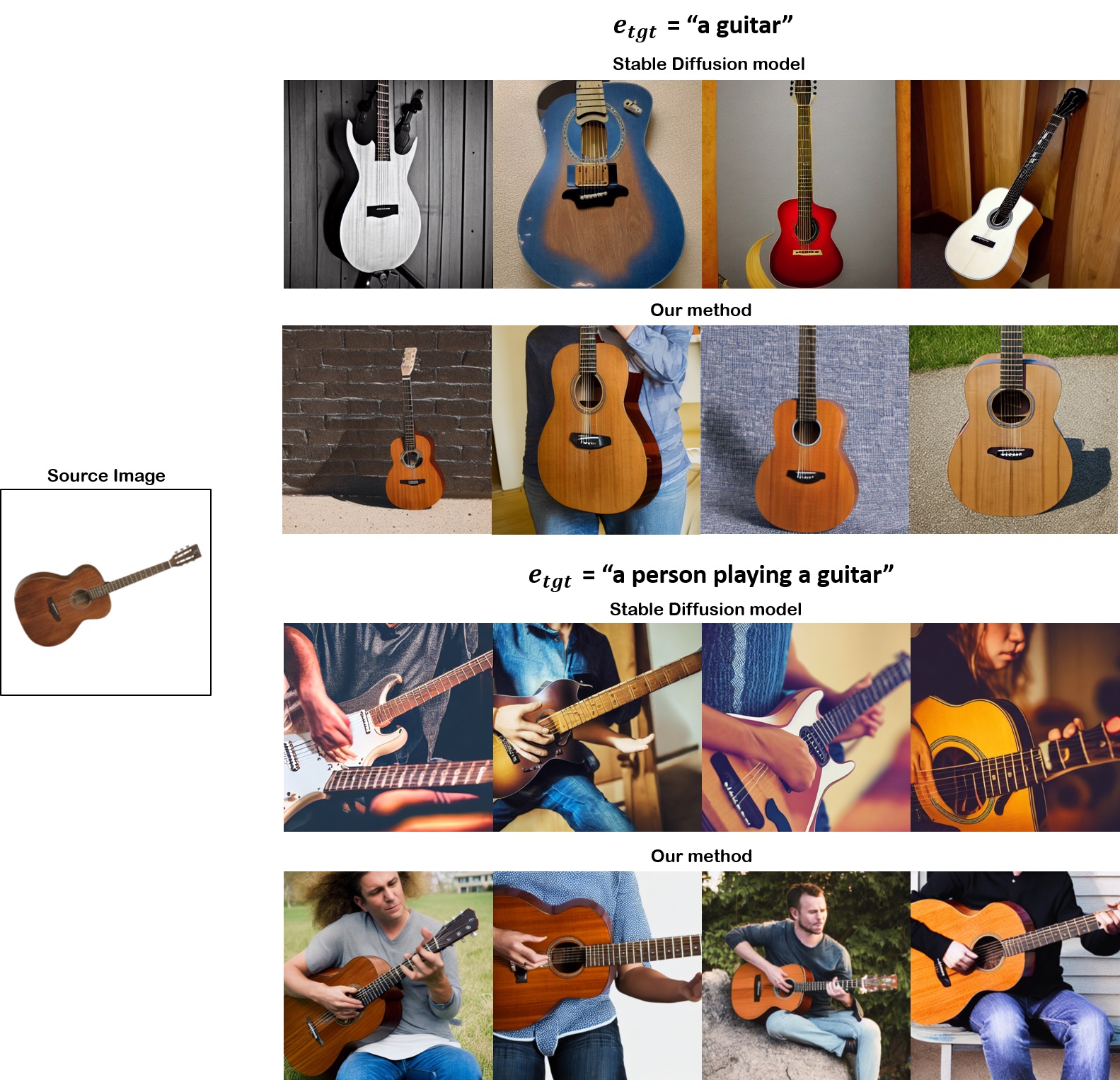}}
    \caption{Ablation study results on the role of $\eb_{hper}$. While the images generated from $\eb_{tgt}$ (i.e. Stable Diffusion) cannot preserve the guitar's identity, guitars in the images generated from $\eb_{hper}$ show same identity as the source image.}
    \vspace{-0.3cm}
    \label{fig_ablation_ehper}
\end{figure}

\begin{figure}[!t]
    \centerline{\includegraphics[width=0.95\columnwidth]{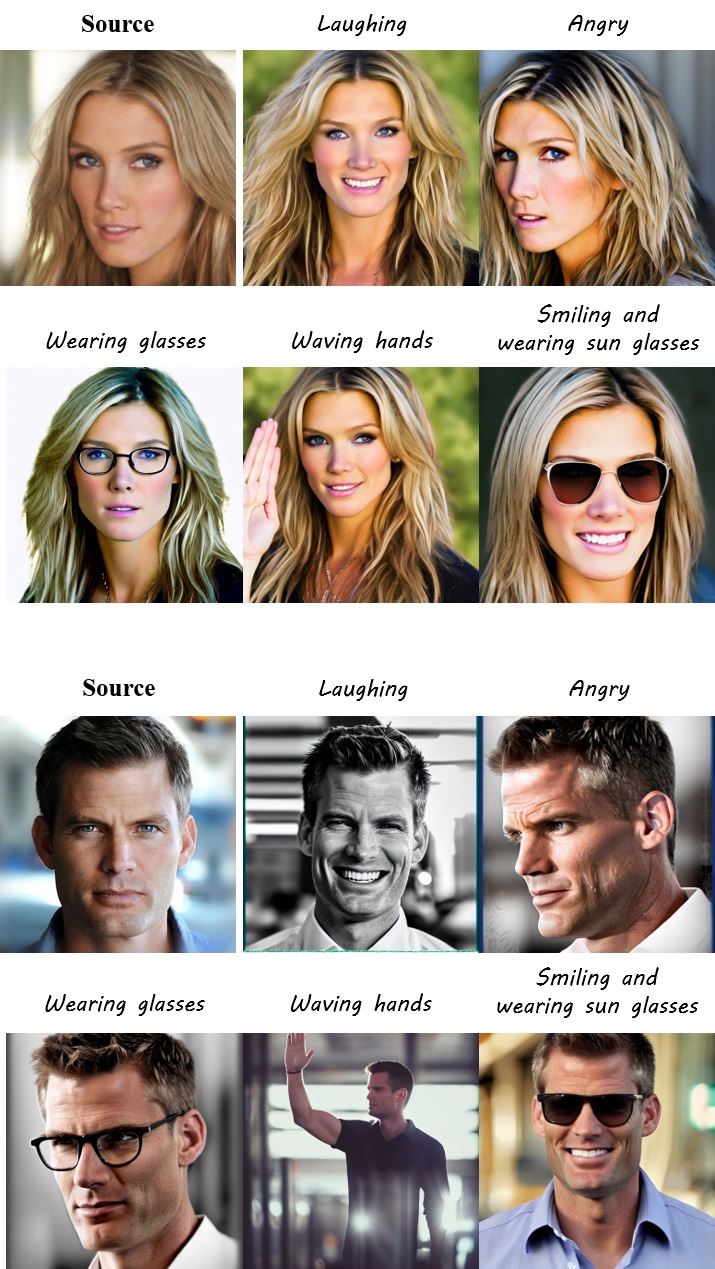}}
    \caption{The facial editing results are presented. As shown, our method is capable of generating images with high-quality facial editing, such as changing the motion, adding glasses, or modifying the facial expression while retaining the identity.}%
    \vspace{0.5cm}
    \label{fig_face_edit}
\end{figure}

\begin{figure}[!t]
    \centerline{\includegraphics[width=0.9\columnwidth]{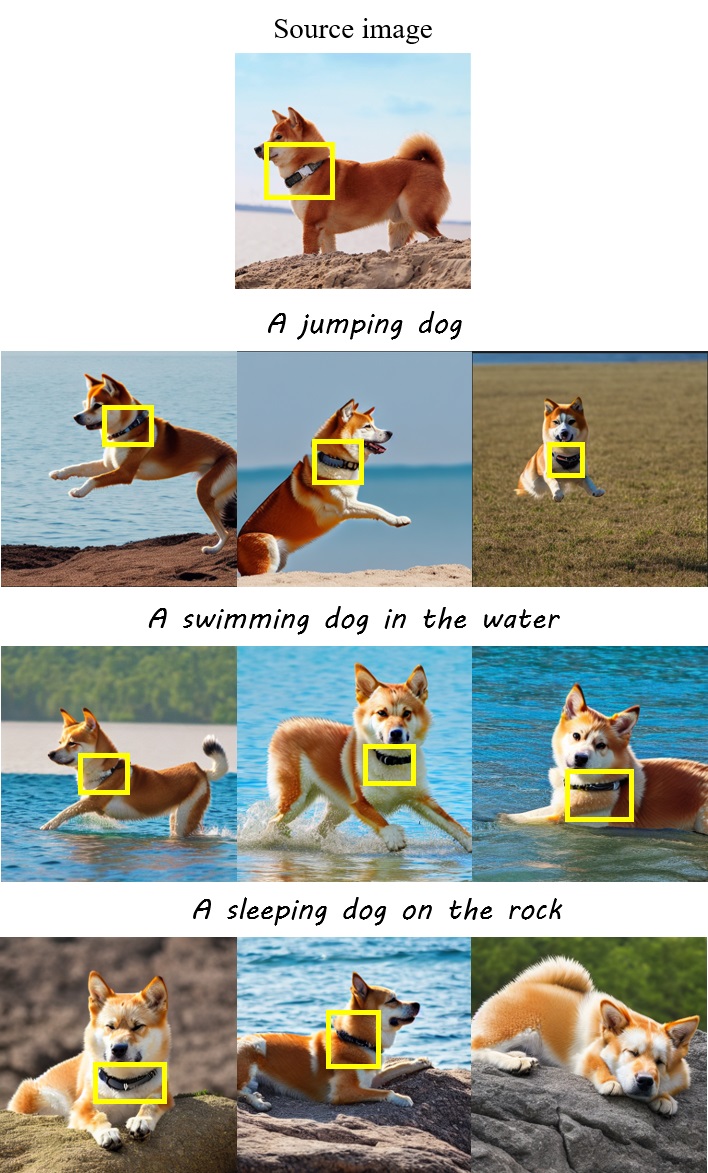}}
    \caption{Personalization results with random seeds. The majority of output images included the dog with its leash. }%
    \vspace{0.5cm}
    \label{fig_seed}
\end{figure}

\begin{figure*}[!t]
    \centerline{\includegraphics[width=1.0\textwidth]{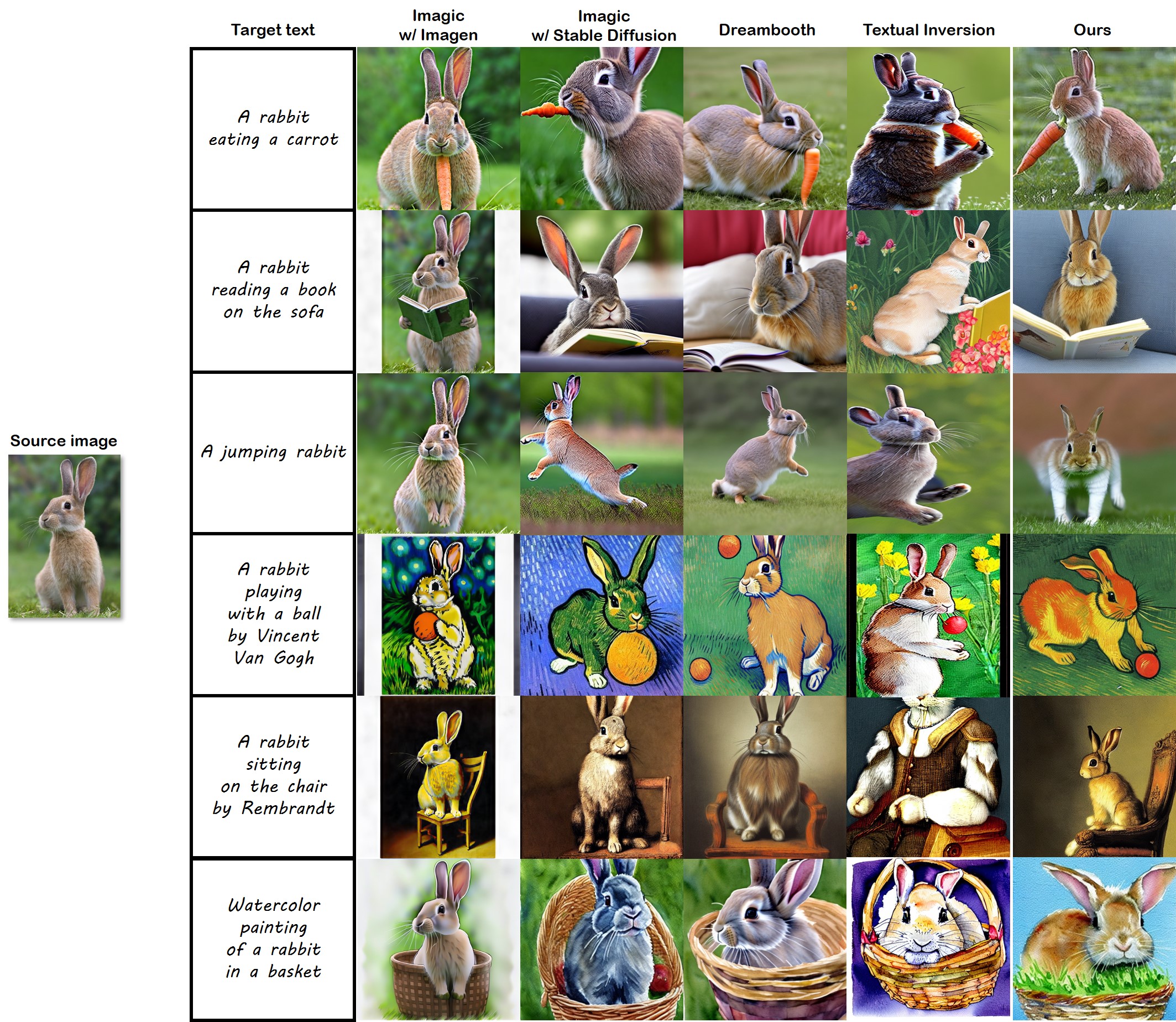}}
    \caption{The qualitative comparison results with one source image are presented. As can be seen, our method consistently produces images that are visually more aligned with the given text prompts and are more similar to the source image in terms of identity. }
    \vspace{-0.3cm}
    \label{rabbit_control}
\end{figure*}

\section{Additional experiments}
\subsection{Ablation study on source embedding $\eb_{src}$}
We conducted an ablation study using different source embeddings to investigate the role of $\eb'_{src}$ during the optimizing phase. As shown in Figure \ref{fig_ablation_esrc}, we found that the identity of the dog in the source image was well-preserved across all source embeddings. However, the output images generated from the prompt ``a standing dog" demonstrated superior performance compared to the other embeddings. In some cases, the dogs in the first row (i.e. ``a dog'') did not resemble the dog in the source image, and the dogs in the third row (i.e. ``a jumping dog'') did not appear to be jumping. Additionally, human-like artifacts were present in the third and last rows. Based on these observations, we decided to use more descriptive source text such as ``a standing dog" for training $\eb_{hper}$ as the baseline.

\subsection{Ablation study on personalized embedding}
In Figure \ref{fig_ablation_ehper}, we examined the function of $\eb_{hper}$. The images generated solely using $\eb_{tgt}$, which corresponds to the Stable Diffusion Model, were not able to preserve the identity of the guitar. On the other hand, the guitars in the images generated from $\eb_{hper}$ retained the same identity as the source image.

\subsection{Additional Results}
Our method using HiPer embedding could manipulate human facial images as illustrated in Figure \ref{fig_face_edit}. Using only a single facial image, we could generate diverse manipulated images. The hair style and facial features of the person in the source image were accurately reflected in the resulting image. Also, it appears that the images are well generated in accordance with the given text prompt.

Furthermore, as demonstrated in Figure \ref{fig_seed}, we achieved high-quality edited images across multiple random seeds while maintaining the identity of the dog and aligning with the target prompts. It is worth noting that the majority of output images included the dog with its leash.

For further comparison, we generated comparative results for additional result images (Figure \ref{sup_control} and \ref{rabbit_control}). From the manipulated images, which are obtained with the Stable Diffusion, it is confirmed that our model produces better results than other models in maintaining the appearance of objects in the source image and reflecting the given text in the resulting image.

\section{Limitations}
\begin{figure}[!t]
    \centerline{\includegraphics[width=1.1\columnwidth]{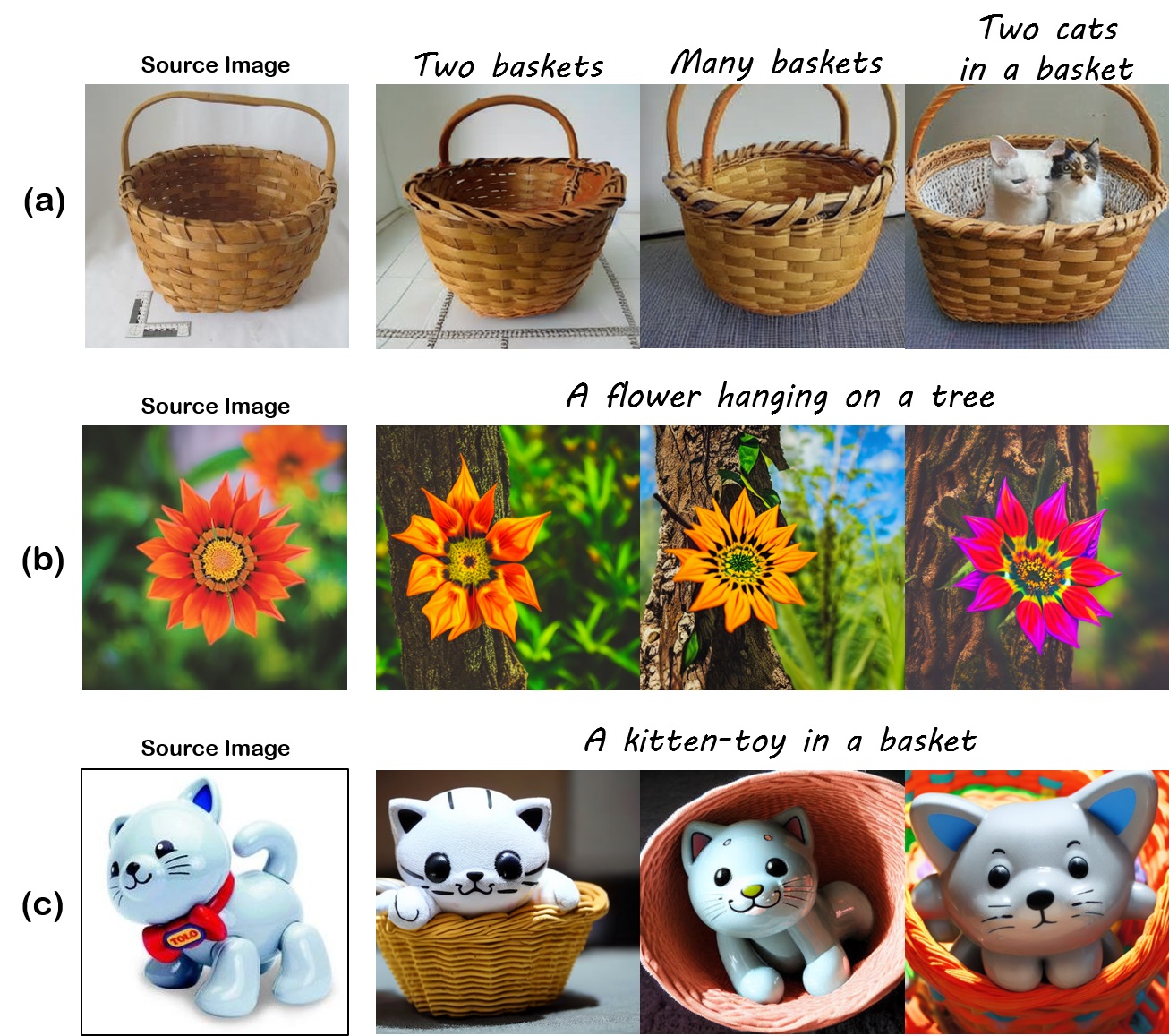}}
    \caption{Failure cases of (a) counting, (b) color, and (c) complex artificial products.}
    \label{fig_failure_total}
\end{figure}

Although HiPer can successfully manipulate images using a wide range of prompts, such as motion, background, and artistic styles, it may encounter difficulties when attempting to manipulate images using certain types of prompts, such as counting (Figure \ref{fig_failure_total}(a)). Although the number of new subjects (``two cats") from the source prompt is incorporated into the output image, the number of the original subject (``two baskets" or ``many baskets") in the source image could not be altered. Also, while the object of generated image with HiPer embedding follows the shape of source image's desired object, there were instances where the color did not match accurately (Figure \ref{fig_failure_total}(b)). If the characteristics of color are not clear or if the color is not specified in the target text prompt, the resulting image may differ in color from the source image.
Another drawback of our approach is that it proved to be ineffective in the production of complex artificial products (Figure \ref{fig_failure_total}(c)). On the other hand, our approach demonstrates superior performance when applied to natural images such as those featuring dogs, birds, or various scenes.

%Sometimes, Stable Diffusion models conditioned on given text prompt generate the image which does not contain the meaning of the given text. Although most personalization of generated image are successful, not all generated images with our method contain successfully the meaning of given text.
%The resulting image retains the shape and characteristics of the source image, but may exhibit changes in color to a different hue.

Our model preserves the identity of the source image, but it does not currently achieve a level where the generated image appears natural. Since our focus is on preserving the identity of the source image, there are unnatural aspects in the outside of the personalization. This is because our image manipulation with the backbone of Stable Diffusion Model, which is publicly available, has the limitation of generating realistic image.

\section{Social impact}
Our method enables manipulation of images using arbitrary prompts, therefore, it is important to exercise caution when selecting images such as those depicting human faces or bodies.
%HiPer is a useful method for manipulating image based on given text. However, 
Moreover, text-to-image manipulation method may have the negative potential to create convincing fake images that could mislead people. Therefore, we recommend that users use the tool appropriately and carefully.

%\section{Code}
%The implementation code for HiPer will be available upon request. 
%%with the following commands:
%\begin{verbatim}
%$ git clone https://
%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@
%github.com/HiPer0/HiPer.git
%\end{verbatim}
%When copying the above commands, be careful with the spacing. To conduct experiments, please install the required packages and do experiments. The detailed information about the implementation can be found in the 'README.md' file.
%
%

\section{Acknowledgements}
We would like to express our gratitude to Shiran Zada and the other authors from Imagic for their help with the official Imagic results.

\section{Source text prompt during training}

\begin{table}[hbt]
\centering
\resizebox{0.3\textwidth}{!}{%
\begin{tabular}{@{}lll@{}}
\toprule
          & \multicolumn{1}{l|}{}        & \multicolumn{1}{c}{source text} \\ \midrule
Figure 1  & \multicolumn{1}{l|}{left}    & a rabbit                        \\
          & \multicolumn{1}{l|}{right}   & a standing dog                  \\ \midrule
Figure 4  & \multicolumn{1}{l|}{1st row} & a bird                          \\
          & \multicolumn{1}{l|}{2nd row} & a standing dog                  \\
          & \multicolumn{1}{l|}{3rd row} & a teddy                         \\
          & \multicolumn{1}{l|}{4th row} & a standing dog                  \\
          & \multicolumn{1}{l|}{5th row} & a bear                          \\ \midrule
Figure 5  & \multicolumn{1}{l|}{1st row} & a car                           \\
          & \multicolumn{1}{l|}{2nd row} & a standing dog                  \\
          & \multicolumn{1}{l|}{3rd row} & a purse                         \\
          & \multicolumn{1}{l|}{4th row} & a cat                           \\ \midrule
Figure 6  & \multicolumn{1}{l|}{}        & a doctor                        \\ \midrule
Figure 7  & \multicolumn{1}{l|}{}        & a standing dog                  \\ \midrule
Figure 8  & \multicolumn{1}{l|}{1st row}        & a cup                           \\ \midrule
Figure 9  & \multicolumn{1}{l|}{}        & a standing dog                  \\ \midrule
Figure 10 & \multicolumn{1}{l|}{}        & a guitar                        \\ \midrule
Figure 11 & \multicolumn{1}{l|}{top}     & a woman                         \\
          & \multicolumn{1}{l|}{bottom}  & a man                           \\ \midrule
Figure 12 & \multicolumn{1}{l|}{}        & a standing dog                  \\ \midrule
Figure 13 & \multicolumn{1}{l|}{}        & a rabbit                        \\ \midrule
Figure 14 & \multicolumn{1}{l|}{1st row} & a basket                        \\
          & \multicolumn{1}{l|}{2nd row} & a flower                        \\
          & \multicolumn{1}{l|}{3rd row} & a kitten-toy                     \\ \midrule
          &                              &                                
\end{tabular}
}
\caption{Source text prompts for generating the figures.}
\label{table_sourcetext}
\end{table}

\end{document}

%% file: packages.tex
\usepackage{multirow}
\usepackage{booktabs}
\usepackage{bm}
\usepackage{multirow}
\usepackage{microtype}
\usepackage{float}
\usepackage{subcaption}

%% file: macros.tex
\usepackage{xkcdcolors}
\newcommand{\xmath}[1] {\ensuremath{#1}\xspace}
\newcommand{\blmath}[1] {\xmath{\bm{#1}}}
\newcommand{\I}{\blmath{I}}
\newcommand{\x}{\blmath{x}}

\newcommand{\Ib}{{\blmath I}}

\newcommand{\eb}{{\blmath e}}

\newcommand{\pb}{{\blmath p}}

\newcommand{\xb}{{\blmath x}}
\newcommand{\yb}{{\blmath y}}

\newcommand{\Nc}{\mathcal{N}}
\newcommand{\Rd}{{\mathbb R}}

\newcommand{\Ed}{{{\mathbb E}}}

\newcommand{\beq}{\begin{equation}}
\newcommand{\eeq}{\end{equation}}
\newcommand{\beqa}{\begin{eqnarray}}
\newcommand{\eeqa}{\end{eqnarray}}

\newcommand{\epsilonb}{\boldsymbol{\epsilon}}

\usepackage{capt-of}
\usepackage{diagbox}